\documentclass[journal]{IEEEtran}
\usepackage{times}

\usepackage[numbers]{natbib}
\usepackage{multicol}
\usepackage[bookmarks=true]{hyperref}

\usepackage{comment}

\usepackage{graphics} % for pdf, bitmapped graphics files
\usepackage{epsfig} % for postscript graphics files
\usepackage{times} % assumes new font selection scheme installed
\usepackage{amsmath} % assumes amsmath package installed
\usepackage{amssymb}  % assumes amsmath package installed
\usepackage{caption}
\usepackage{subcaption}
\let\labelindent\relax
\usepackage[shortlabels]{enumitem}
\usepackage{multirow}
\usepackage{booktabs}
\usepackage[nocomma]{optidef}
\usepackage{cancel}
\usepackage{color}
\usepackage[normalem]{ulem}

\DeclarePairedDelimiter\abs{\lvert}{\rvert}%
\DeclarePairedDelimiter\norm{\lVert}{\rVert}%

\makeatletter
\let\oldabs\abs
\def\abs{\@ifstar{\oldabs}{\oldabs*}}
\let\oldnorm\norm
\def\norm{\@ifstar{\oldnorm}{\oldnorm*}}
\makeatother

\newcommand{\ecmnnlong}{Equality Constraint Manifold Neural Network}
\newcommand{\ecmnn}{ECoMaNN}
\newcommand{\configspace}{\mathcal{C}}
\newcommand{\dimambient}{d}
\newcommand{\dimconstraint}{l}

\newcommand{\jacobian}{\mathbf{J}}
\newcommand{\jointposition}{\mathbf{q}}

\newcommand{\numconstraintmanifold}{n}
\newcommand{\constraintmanifold}{M}
\newcommand{\constraintmanifoldsequence}{\mathcal{\constraintmanifold}}
\newcommand{\constraintfunction}{{h}_{\constraintmanifold}}
\newcommand{\constraintmanifoldjacobian}{\jacobian_\constraintmanifold}
\newcommand{\onconstraintconfigspace}{{\configspace}_{\constraintmanifold}}
\newcommand{\offconstraintconfigspace}{{\configspace}_{\cancel{\constraintmanifold}}}
\newcommand{\numnearestneighbor}{K}
\newcommand{\idxnearestneighbor}{k}
\newcommand{\KNNset}{\mathcal{K}}
\newcommand{\origKNN}{\hat{\KNNset}}
\newcommand{\recenteredKNN}{\tilde{\KNNset}}
\newcommand{\nearestneighborjointposition}{\hat{\jointposition}}
\newcommand{\recenterednearestneighborjointposition}{\tilde{\jointposition}}
\newcommand{\designmatrix}{\mathbf{X}}
\newcommand{\samplecovariancematrix}{\mathbf{S}}

\newcommand{\diagsingularvalues}{\mathbf{\Sigma}}

\newcommand{\righteigmat}{\mathbf{V}}
\newcommand{\covdiagsingularvalues}{\diagsingularvalues}

\newcommand{\covrighteigmat}{\righteigmat}
\newcommand{\lpcacoordframe}{\mathcal{F}}
\newcommand{\coveigvec}{\vct{v}}
\newcommand{\constrjaceigvec}{\vct{e}}
\newcommand{\nullspaceid}{\text{N}}
\newcommand{\covnullspaceeigmat}{\righteigmat_{\nullspaceid}}
\newcommand{\constrjacrighteigmat}{\mathbf{E}}
\newcommand{\constrjacnullspaceeigmat}{\constrjacrighteigmat_{\nullspaceid}}
\newcommand{\augjointposition}{\check{\jointposition}}
\newcommand{\augidx}{i}
\newcommand{\augmagnitude}{\epsilon}
\newcommand{\cost}{J}
\newcommand{\tpath}{\tau}
\newcommand{\tpathset}{\vct{\tpath}}
\newcommand{\costfunctional}{\mathcal{\cost}}
\newcommand{\tpathspace}{\mathcal{T}}
\newcommand{\freeconfigspaceoperator}{\Upsilon}

\newcommand{\argmin}{\operatorname*{arg\:min}}

\renewcommand{\Re}{\mathbb{R}}

\newcommand{\vct}[1]{\boldsymbol{#1}} % vector
\newcommand{\T}{^{\textrm T}} % transpose

\newcommand{\period}{~.}
\renewcommand{\t}[1]{{\textrm{#1}}}
\newcommand{\st}{\t{s.t.}}

\begin{document}

% paper title
\title{Learning Manifolds for Sequential Motion Planning}

% You will get a Paper-ID when submitting a pdf file to the conference system
% \author{Author Names Omitted for Anonymous Review. Paper-ID [add your ID here]}

% \author{
% \authorblockN{Isabel M. Rayas Fern{\'a}ndez\authorrefmark{1}, 
% Giovanni Sutanto\authorrefmark{1}, 
% Peter Englert, 
% Ragesh K. Ramachandran, and 
% Gaurav S. Sukhatme}
% \authorblockA{Robotic Embedded Systems Laboratory\\
% University of Southern California, Los Angeles CA}
% \authorblockA{\authorrefmark{1}equal contribution}
% }

\author{
Isabel M. Rayas Fern{\'a}ndez\authorrefmark{1}, 
Giovanni Sutanto\authorrefmark{1}, 
Peter Englert, 
Ragesh K. Ramachandran, 
Gaurav S. Sukhatme%
\thanks{\authorrefmark{1} equal contribution}%
\thanks{All authors are with the Robotic Embedded Systems Laboratory,
University of Southern California, Los Angeles, CA, USA.}%
\thanks{This material is based upon work supported by the National Science Foundation Graduate Research Fellowship Program under Grant No. DGE-1842487. Any opinions, findings, and conclusions or recommendations expressed in this material are those of the author(s) and do not necessarily reflect the views of the National Science Foundation.
This work was supported in part by the Office of Naval Research (ONR) under grant N000141512550.}
}

\maketitle

\begin{abstract}
% We expand on the previous work in constrained robot motion planning on manifold sequences.
% We consider the case where a manifold constraint is not known analytically.
% Instead, we have used techniques from machine learning to learn representations of the constraints from data. 
% We compare several approaches for multiple robotic datasets,
% and we discuss how to incorporate the learned constraints into an existing planning framework.
% We conclude with insights learned and next steps.
Motion planning with constraints is an important part of many real-world robotic systems. 
In this work, we study manifold learning methods to learn such constraints from data.
We explore two methods for learning implicit constraint manifolds from data: Variational Autoencoders (VAE), and a new method, {\ecmnnlong} ({\ecmnn}). With the aim of incorporating learned constraints into a sampling-based motion planning framework, we evaluate the approaches on their ability to learn representations of constraints from various datasets and on the quality of paths produced during planning.
\end{abstract}

\IEEEpeerreviewmaketitle

\section{Introduction}
\label{sec:introduction}

% introduce topic of sequential motion planning in TAMP 
Many practical robotic applications require planning robot motions with constraints, such as maintaining an orientation or reaching a particular location. Planning becomes more complicated when the  task consists of many subtasks that must be completed in sequence. In this case, task and motion planning frameworks \cite{kaelbling2013integrated, dantam2016incremental, konidaris2018skills, toussaint2018differentiable, dantam2018task, barry2013hierarchical, cambon2009hybrid} can be used to handle long planning horizons and a wide range of tasks. 
% motivation: some constraints are hard to describe; some maybe hard to sample from; we show this for a wide variety of constraint types; give pouring example
However, some constraints may be difficult to describe analytically, or it may be difficult to sample constraints that adhere to them. For example, if the task is to pour a liquid from a bottle into a cup, it is not immediately clear how to encode the motion constraints for a planning algorithm.

% describe our work at a high level
In this work, we focus on learning constraint manifolds for use in constrained motion planning algorithms \cite{englert2020sampling, stilman2010global, berenson2011task, jaillet2013asymptotically, kim2016tangent, kingston2019ijrr, csucan2012motion, cortes2004sampling}.
To this end, we investigate two different approaches: Variational autoencoders (VAE) following \citet{kingma2013AutoEncodingVB}, and {\ecmnnlong}  (\ecmnn), a method we propose which learns the implicit function value of equality constraints. We evaluate these techniques on six datasets of varying size and complexity, and we present preliminary resulting motion plans.

\section{Background on Sequential Manifold Planning}
\label{sec:background}
We focus on learning manifolds that describe kinematic robot tasks. We aim to integrate these learned manifolds into the sequential manifold planning (SMP) framework proposed in \citet{englert2020sampling}.
SMP considers kinematic motion planning problems in a configuration space $\configspace \subseteq \Re^{\dimambient}$. A robot configuration 
% $q\in C$ 
$\jointposition \in \configspace$
describes the state of one or more robots with 
% $k$ 
$\dimambient$
degrees of freedom in total.
A manifold 
% $M$ 
$\constraintmanifold$
is represented as an equality constraint 
% $h_{M}(q)=0$ 
$\constraintfunction(\jointposition) = \vct{0}$
where 
% $h_{M}(q) : \R^{k} \to \R^{l}$. 
$\constraintfunction: \Re^{\dimambient} \rightarrow \Re^{\dimconstraint}$
and $\dimconstraint$ is the dimensionality of the implicit manifold.
The set of robot configurations that are on a manifold 
% $M$ 
$\constraintmanifold$
is given by 
% $$C_{M} = \{q\in C ~|~ h_{M}(q) = 0\}\period$$
\mbox{$\onconstraintconfigspace = \{ \jointposition \in \configspace \mid \constraintfunction(\jointposition) = \vct{0} \}\period$}
SMP defines the planning problem as a sequence of 
% $n+1$ 
$(\numconstraintmanifold+1)$
such manifolds 
% $$\mathcal{M} = \{M_1, M_2, \dots, M_{n+1}\}$$
\mbox{$\mathcal{\constraintmanifold} = \{\constraintmanifold_1, \constraintmanifold_2, \dots, \constraintmanifold_{\numconstraintmanifold+1}\}$}
and an initial configuration 
% $q_\t{start}\in C_{M_1}$ 
$\jointposition_\t{start} \in \configspace_{{\constraintmanifold}_1}$ 
that is on the first manifold.
The goal is to find a path from 
% $q_\t{start}$ 
$\jointposition_\t{start}$ 
that traverses the manifold sequence 
% $\mathcal{M}$ 
$\constraintmanifoldsequence$
and reaches a configuration on the goal manifold 
% $M_{n+1}$. 
$\constraintmanifold_{\numconstraintmanifold+1}$. 
A path on the $i$-th manifold is defined as 
% $\tau_i : [0, 1] \to C_{M_i}$
$\tpath_i : [0, 1] \to {\onconstraintconfigspace}_{i}$
and 
% $J(\tau_i)$ 
$\cost(\tpath_i)$ 
is the cost function of a path
% $\mathcal{J} : \mathcal{T} \to \R_{\geq 0}$
$\costfunctional : \tpathspace \to \Re_{\geq 0}$
where 
% $\mathcal{T}$ 
$\tpathspace$ 
is the set of all non-trivial paths. The problem is formulated as an optimization over a set of paths 
% $\vct\tau = (\tau_1, \dots, \tau_{n})$ 
$\tpathset = (\tpath_1, \dots, \tpath_{\numconstraintmanifold})$ 
that minimizes the sum of path costs under the constraints of traversing 
% $\mathcal{M}$ 
$\constraintmanifoldsequence$ 
and of being collision-free:
% \begin{align} %
% 	\begin{alignedat}{2} %	
% 	\label{eq:smp_problem}
% 	&\vct\tau^{\star} = \argmin_{\vct\tau} \sum_{i=1}^n J(\tau_i) &&\\
% 	\st\quad &\tau_1 (0) = q_\t{start} &\\
% 	&\tau_i(1) = \tau_{i+1}(0)  &&\forall_{i=1,\dots,n-1} \\ 
% 	&C_{\t{free}, i+1} = \Upsilon(C_{\t{free}, i}, \tau_{i})~~~~ &&\forall_{i=1,\dots,n}\\
% 	&\tau_i(s) \in C_{M_i} \cap C_{\t{free}, i} &&\forall_{i=1,\dots,n}~ \forall_{s \in [0, 1]}\\
% 	&\tau_{n}(1) \in C_{M_{n+1}} \cap C_{\t{free}, n+1} &&
% 	\end{alignedat} %
% \end{align}
\begin{align} %
	\begin{alignedat}{2} %	
	\label{eq:smp_problem}
	&\tpathset^{\star} = \argmin_{\tpathset} \sum_{i=1}^{\numconstraintmanifold} \cost(\tpath_i) &&\\
	\st\quad &{\tpath}_1 (0) = {\jointposition}_\t{start} &\\
	&\tpath_i(1) = \tpath_{i+1}(0)  &&\forall_{i=1,\dots,\numconstraintmanifold-1} \\ 
	&\configspace_{\t{free}, i+1} = \freeconfigspaceoperator(\configspace_{\t{free}, i}, \tpath_{i})~~~~ &&\forall_{i=1,\dots,\numconstraintmanifold}\\
	&\tpath_i(s) \in {\onconstraintconfigspace}_{i} \cap \configspace_{\t{free}, i} &&\forall_{i=1,\dots,\numconstraintmanifold}~ \forall_{s \in [0, 1]}\\
	&\tpath_{\numconstraintmanifold}(1) \in {\onconstraintconfigspace}_{\numconstraintmanifold+1} \cap \configspace_{\t{free}, \numconstraintmanifold+1} &&
	\end{alignedat} %
\end{align}
%
% $\Upsilon$ 
$\freeconfigspaceoperator$
is an operator that describes the change in the free configuration space (the space of all configurations that are not in collision with the environment)
% $C_\t{free}$ 
$\configspace_\t{free}$ 
when transitioning to the next manifold. The SMP algorithm is able to solve this problem for a certain class of problems. It iteratively applies RRT${}^\star$ to find a path that reaches the next manifold while staying on the current manifold.
For further details of the SMP algorithm, we refer the reader to \citet{englert2020sampling}. 

In this paper, we employ data-driven algorithms to learn manifolds $M$ from data with the goal to integrate them into the SMP framework.
%
% maybe we don't need this section, if we beef up the intro paragraph a little
% or, maybe we include this but just explain how manifolds were represented before.
% TOOD: need to define h(q), C_M

\section{Manifold Learning}
\label{sec:manifold_learning}
Learning constraint manifolds from data is attractive for multiple reasons. 
For example, it may be easier for a human to demonstrate a task rather than specifying constraints analytically,
% we may want to encode constraints that are not strict equalities, 
or we may want to reduce the amount of expert information needed. 

We propose a novel neural network structure -- called \emph{\ecmnnlong} (\ecmnn) -- to become a learning representation that takes $\jointposition$ as input and outputs the prediction of the implicit function $\constraintfunction(\jointposition)$.
Moreover, we would like to train {\ecmnn} in a supervised manner, from demonstrations. One of the challenges is that the supervised training dataset is collected \emph{only} from demonstrations of data points which are on the equality constraint manifold $\onconstraintconfigspace$, called the \emph{on-manifold} dataset.
% and potentially corrupted with some noise. 
This is a reasonable assumption, since collecting both the on-manifold $\onconstraintconfigspace$ and off-manifold $\offconstraintconfigspace = \{ \jointposition \in \configspace \mid \constraintfunction(\jointposition) \neq \vct{0} \}$ datasets for supervised training will be tedious because the implicit function $\constraintfunction$ values of points in $\offconstraintconfigspace$ are typically unknown and hard to label.
We will show that even though our approach is only provided with data on $\onconstraintconfigspace$, it can still learn a useful representation of the manifold, sufficient for use in the SMP framework.

% I am not sure how to say this... Is this local neighborhood called a chart in Differential Geometry? Maybe Ragesh can help a bit?
Our goal is to learn a single global representation of the constraint manifold in form of a neural network. 
% However, according to Differential Geometry \citep{Lee00_IntroToSmoothManifolds}, each local neighborhood on the manifold contains important information. Moreover, collectively all these local neighborhoods characterize the global representation of the manifold. 
A manifold can be defined as a collection of local neighborhoods 
% called \emph{charts}
which resemble Euclidean spaces (\citet{Lee00_IntroToSmoothManifolds}). Therefore, a global representation of the manifold can be developed by constructing characterizations for its Euclidean-like local neighborhoods.

Our approach leverages local information on the manifold in the form of the tangent and normal spaces (\citet{Deutsch2015_TensorVotingGraph, GStrangIntroLinearAlgebra}).
% For a point on the manifold, the tangent and normal spaces are valid only for a small neighborhood surrounding the point. 
% With regard to the implicit function $\constraintfunction$, the tangent space is a vector space from which a vector can be chosen such that following this vector will keep the valuation of the function $\constraintfunction$ unchanged.  \imr{only locally/infintesimally, right? the value will change far from the point where it is tangent} 
% On the other hand, the normal space is a vector space from which a vector can be chosen such that following this vector will result in the maximum change in the valuation of the function $\constraintfunction$. \rag{I think previous two sentences are not required}
With regard to $\constraintfunction$, the tangent and normal spaces are equivalent to the null and row space,
% \rag{i think this should be row space or coimage space} 
respectively, of the matrix 
% $\constraintmanifoldjacobian = \frac{\partial \constraintfunction}{\partial \jointposition}$ 
$\constraintmanifoldjacobian = \left. \frac{\partial \constraintfunction({\jointposition})}{\partial {\jointposition}}\right|_{{\jointposition} = \Bar{\jointposition}}$, 
% evaluated at a point $\jointposition$ 
% \rag{$\Bar{\jointposition} = \jointposition$}, 
and valid in a small neighborhood around the point $\Bar{\jointposition}$.

Using on-manifold data, the local information of the manifold can be analyzed using Local Principal Component Analysis (Local PCA) (\citet{Kambhatla_LocalPCA}). Essentially, for each data point $\jointposition$ in the on-manifold dataset, we establish a local neighborhood using $\numnearestneighbor$-nearest neighbors ($\numnearestneighbor$NN) $\origKNN = \{\nearestneighborjointposition_1, \nearestneighborjointposition_2, \dots \nearestneighborjointposition_\numnearestneighbor\}$, with $\numnearestneighbor \geq \dimambient$. 
After a change of coordinates, $\jointposition$ becomes the origin of a new local coordinate frame $\lpcacoordframe$, and the $\numnearestneighbor$NN becomes $\recenteredKNN = \{\recenterednearestneighborjointposition_1, \recenterednearestneighborjointposition_2, \dots \recenterednearestneighborjointposition_\numnearestneighbor\}$ with $\recenterednearestneighborjointposition_\idxnearestneighbor = \nearestneighborjointposition_\idxnearestneighbor - \jointposition$ for all values of $\idxnearestneighbor$. Defining the matrix 
$\designmatrix = 
\begin{bmatrix} 
\recenterednearestneighborjointposition_1 & \recenterednearestneighborjointposition_2 & \hdots & \recenterednearestneighborjointposition_\numnearestneighbor \\
\end{bmatrix}\T \in \Re^{\numnearestneighbor \times \dimambient}
$, we can compute the covariance matrix $\samplecovariancematrix = \frac{1}{\numnearestneighbor-1} \designmatrix\T \designmatrix \in \Re^{\dimambient \times \dimambient}$.
The eigendecomposition of $\samplecovariancematrix = \covrighteigmat \covdiagsingularvalues \covrighteigmat\T$ gives us the Local PCA. 
The matrix $\covrighteigmat$ contains the eigenvectors of $\samplecovariancematrix$ as its columns in decreasing order w.r.t.\ the corresponding eigenvalues in the diagonal matrix $\covdiagsingularvalues$. These eigenvectors form the basis of $\lpcacoordframe$.

This local coordinate frame $\lpcacoordframe$ is tightly related to the tangent and normal spaces of the manifold at $\jointposition$. That is, the $(\dimambient - \dimconstraint)$ eigenvectors corresponding to the $(\dimambient - \dimconstraint)$ biggest eigenvalues of $\covdiagsingularvalues$ form a basis of the tangent space, while the remaining $\dimconstraint$ eigenvectors form the basis of the normal space. 
Furthermore, due to the characteristics of the manifold from which the dataset was collected, the $\dimconstraint$ smallest eigenvalues of $\covdiagsingularvalues$ will be close to zero, resulting in the $\dimconstraint$ eigenvectors associated with them forming the basis of the null space of $\samplecovariancematrix$. On the other hand, the remaining $(\dimambient - \dimconstraint)$ eigenvectors form the basis of the row space of $\samplecovariancematrix$. 
% Thus at $\jointposition$, the tangent space is equivalent to both the null space of $\constraintmanifoldjacobian$ and the row space of $\samplecovariancematrix$, while the normal space is equivalent to the row space of $\constraintmanifoldjacobian$ and the null space of $\samplecovariancematrix$.

To this end, we present several methods to define and train \ecmnn, as follows:
% \subsubsection{Determination of the Number of Constraints}
% ~\\
% We follow the same technique in \citep{Deutsch2015_TensorVotingGraph} for automatically determining the number of constraints $\dimconstraint$ from data, which is also the number of outputs of {\ecmnn}\footnote{Here we assume that the intrinsic dimensionality of the manifold at each point remains constant.}. Suppose the eigenvalues of $\samplecovariancematrix$ is $\{\covsingularvalue_1, \covsingularvalue_2, \dots, \covsingularvalue_\dimambient\}$ (in a decreasing order w.r.t. magnitude), then the number of constraint can be determined as $\dimconstraint = \argmax{\left(\left[\coveigval_1 - \coveigval_2, \coveigval_2 - \coveigval_3, \dots, \coveigval_{\dimambient-1} - \coveigval_{\dimambient}
% \right]\right)}$.

\subsection{Local Tangent and Normal Spaces Alignment}
% ~\\
% Previously Local Tangent Space Alignment (LTSA) \citep{Zhang02_LTSA} is a well-known manifold learning technique for embedding high-dimensional data into low-dimensional data, which is done by aligning the tangent space of each local neighborhood one another across the manifold. On the other hand, our method does not aim to achieve a low-dimensional embedding. Instead we try to learn the high-dimensional constraint manifold representation of the data for later use in motion planning.
% To-Do: explain why learning the the high-dimensional manifold representation?
% Our approach aims to align the following:
{\ecmnn} aims to align the following:
\begin{enumerate}[(a)]
    \item the null space of $\constraintmanifoldjacobian$ and the row space of $\samplecovariancematrix$, which both must be equivalent to the tangent space, and
    \item the row space of $\constraintmanifoldjacobian$ and the null space of $\samplecovariancematrix$, which both must be equivalent to the normal space
\end{enumerate}
for each local neighborhood of each point $\jointposition$ in the on-manifold dataset. 
Suppose the eigenvectors of $\samplecovariancematrix$ are $\{\coveigvec_1, \coveigvec_2, \dots, \coveigvec_\dimambient\}$ and the singular vectors of $\constraintmanifoldjacobian$ are $\{\constrjaceigvec_1, \constrjaceigvec_2, \dots, \constrjaceigvec_\dimambient\}$, where the indices indicate the decreasing order w.r.t. the eigenvalue/singular value magnitude. The null spaces of $\samplecovariancematrix$ and $\constraintmanifoldjacobian$ are spanned by $\{\coveigvec_{\dimambient-\dimconstraint+1}, \dots, \coveigvec_\dimambient\}$ and $\{\constrjaceigvec_{\dimconstraint+1}, \dots, \constrjaceigvec_\dimambient\}$, respectively. The two conditions above imply that the projection of the null space eigenvectors of $\constraintmanifoldjacobian$ into the null space of $\samplecovariancematrix$ should be $\vct{0}$, and similarly for the row spaces. 
% Similarly, the projection of the null space eigenvectors of $\samplecovariancematrix$ into the null space of $\constraintmanifoldjacobian$ shall be $\vct{0}$. 
Hence, we achieve this by training {\ecmnn} to minimize projection errors $\norm{\covnullspaceeigmat \covnullspaceeigmat\T \constrjacnullspaceeigmat}_2^2$ and $\norm{ \constrjacnullspaceeigmat \constrjacnullspaceeigmat\T \covnullspaceeigmat}_2^2$ with 
$\covnullspaceeigmat = 
\begin{bmatrix}
    \coveigvec_{\dimambient-\dimconstraint+1} & \dots & \coveigvec_\dimambient
\end{bmatrix}$
and 
$\constrjacnullspaceeigmat =  
\begin{bmatrix}
    \constrjaceigvec_{\dimconstraint+1} & \dots & \constrjaceigvec_\dimambient
\end{bmatrix}$.
%
% To-Do: also point out that the eigenvectors of Local PCA does NOT encode orientation, since sample covariance matrix is positive semi-definite, and hence the SVD has the left and right eigenvectors being the same --flipping the polarity of one of the eigenvectors will still lead to the same sample covariance matrix (SVD eigenvector v and -v is the same thing)--. On the other hand, the eigen decomposition of the Jacobian matrix results in eigenvectors who has polarity information, i.e. with eigenvalue fixed and positive, the eigenvector v and -v is NOT the same.
% To-Do: also point out that SVD result (in terms of eigenvectors ordering) for Local PCA is sensitive to the K-nearest-neighbors involved. For example on the case of 3D line constraint, the dimensionality of the constraints is 2, and neither of the eigenvectors in the normal space are dominant; however, a slight bit of noise in the data involved nearby point q1 and q2 (both on the manifold, and q1 != q2), can lead to two different eigenvector decompositions by SVD at q1 and q2. Hence, we cannot align each individual eigenvector of Local PCA with an eigenvector of the Jacobian J, as they may not correspond one another; what we can do is to align the null space of the Local PCA with the row space of the Jacobian, and to align the row space of the Jacobian with the null space of Local PCA, e.g. using projection error minimization.

\subsection{Data Augmentation with Off-Manifold Data}
% ~\\
The training dataset is on-manifold, i.e., each point $\jointposition$ in the dataset satisfies $\constraintfunction(\jointposition) = \vct{0}$. 
Through Local PCA on each of these points, we know the data-driven approximation of the normal space of the manifold at $\jointposition$. 
Hence, we know the directions where the violation of the equality constraint increases, i.e., the same or opposite direction of any vector from the approximate normal space. 
Since our future use of the learned constraint manifold on motion planning does not require the acquisition of the near-ground-truth value of $\constraintfunction(\jointposition) \neq \vct{0}$, we can set this off-manifold valuation of $\constraintfunction$ arbitrarily, as long as it does not interfere with the utility for projecting an off-manifold point onto the manifold. 
Therefore, we can augment our dataset with additional off-manifold data to achieve a more robust learning of {\ecmnn}. 
For each point $\jointposition$ in the on-manifold dataset, and for each eigenvector $\coveigvec$ that forms the basis of the normal space at $\jointposition$, we can add an off-manifold point $\augjointposition = \jointposition + \augidx \augmagnitude \coveigvec$ with a non-zero signed integer $\augidx$ and 
% a small positive scalar $\augmagnitude$ proportional to the square root of the biggest eigenvalue of $\samplecovariancematrix$. 
a small positive scalar $\augmagnitude$.
For such an augmented data point $\augjointposition$, we set the label satisfying $\norm{\constraintfunction(\augjointposition)}_2 = \abs{\augidx} \augmagnitude$. During training, we minimize the prediction error $\norm{(\norm{\constraintfunction(\augjointposition)}_2 - \abs{\augidx} \augmagnitude)}_2^2$ for each augmented point $\augjointposition$.

% To-Do: explain that {\ecmnn} is a generalization of the Signed Distance Function (SDF) Neural Network, up to a scale.

% To-Do: write a pseudo-algorithm describing our approach.

% \input{related_work}
\section{Datasets}
\label{sec:datasets}
We use a robot simulator (\citet{todorov2012mujoco}) to generate various datasets. For each dataset, we define $\constraintfunction(\jointposition)$ by hand and randomly sample points in the configuration space and project them onto the manifold. We use six datasets:
\begin{itemize}
    \item \textbf{Nav}: 2D point that has to stay close to a reference point. Defined as an inequality constraint. $N=15000$.
    \item \textbf{Sphere}: 3D point that has to stay on the surface of a sphere. $N=10000$.
    % renaming all 3DOF to "Plane" 
    \item \textbf{Plane}: Robot arm with 3 rotational DOFs where the end effector has to be on a plane. $N=999$.
    % renaming all 6DOF to "Orient"
    \item \textbf{Orient}: Robot arm with 6 rotational DOFs that has to keep its orientation upright (e.g., transporting a cup). $N=21153$.
    \item \textbf{Tilt}: Same as Orient, but here the orientation constraint is relaxed to an inequality constraint. $N=2000$.
    \item \textbf{Handover}: Robot arm with 6 rotational DOFs and a mobile base with 2 translational DOFs. The manifold is defined as an equality constraint that describes the handover of an object between the two robots. $N=2002$.
\end{itemize}

\section{Experiments}
\label{sec:experiments}
We compare the proposed {\ecmnn} method to a Variational Autoencoder (VAE), which is a popular method for learning a generative
model of a set of data (\citet{chen2016dynamic, kingma2013AutoEncodingVB, park2018multimodal}). Importantly, because
they embed data points as a distribution in the latent space, new latent vectors
can be sampled and decoded into unseen examples which fit the distribution of
the training data. VAEs make use of two neural networks in a neural autoencoder
structure during training, and they only use the decoder during generation. The
key idea that makes VAEs computationally tractable is that the distribution in
the latent space is assumed to be Gaussian. The loss function is a combination of the reconstruction error of the input and the KL divergence of the latent space distribution, weighted by a parameter $\beta$. 

We use the following network structures and parameters: 
For the Nav dataset, the VAE has two hidden layers with 6 and 4 units. The input size is 2 and the embedding size is 2. 
For the Plane dataset, the VAE has three hidden layers with 12, 9, and 6 units. The input size is 3 and the embedding size is 2.
For the Sphere, Orient, Tilt, and Handover datasets, the VAEs have the same structure: Four hidden layers with 8, 6, 6, and 4 units. The input sizes to the networks are 3, 6, 6, and 8, and the embedding sizes are 2, 5, 3, and 7, respectively.
All VAE models have $\beta$ = 0.25 and use batch normalization. We train for 500 epochs for Handover, and 200 otherwise.

% 1. compare h values
\subsection{Evaluate Implicit Functions on Datasets}
\label{sec:experiment1}
We compare the performance of the models using the implicit function value $\constraintfunction$. In the case of the VAE models, we take the reprojected data $\hat{X}$ and evaluate each configuration with $\constraintfunction$. In the case of the \ecmnn, the output of the network is the estimated implicit function value of the input, so we can directly use it. We report the mean and standard deviation of $\constraintfunction$ for each dataset in Table \ref{table:experiment1}. Note that for Nav and Tilt datasets, $h_M$ does not need to be 0 for a configuration to be valid, since these are inequality constraints. Values less than 1 for Nav and less than 0.1 for Tilt adhere to the constraints.

\begin{figure}
    \centering
    \begin{subfigure}[b]{0.66\columnwidth}
        \centering
        \includegraphics[width=\columnwidth]{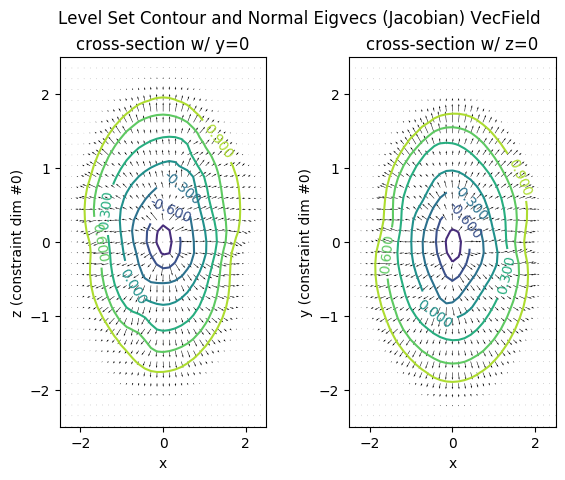}
        \vspace{-0.8cm}
        \caption{Level set contour plot and the learned {\ecmnn}'s normal space eigenvector field, after training on a 3D unit sphere constraint dataset.}
        \label{sfig:levelset_contour_and_jac_vector_field}
    \end{subfigure}
    \hfill
    \begin{subfigure}[b]{0.3\columnwidth}
        \centering
        \includegraphics[width=\columnwidth]{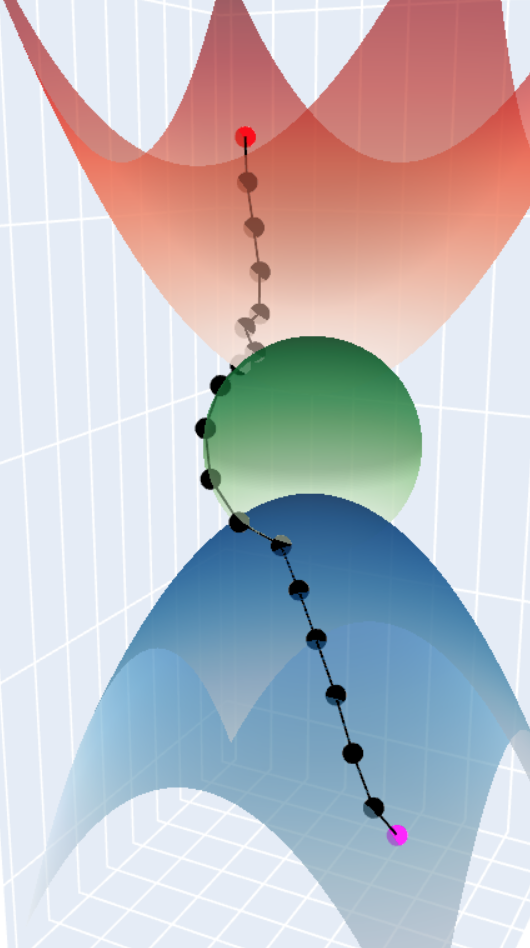}
        \caption{Visualization of a planned path on a learned manifold (sphere).}
        \label{sfig:hourglass_learned}
    \end{subfigure}
\end{figure}

% gsutanto: please feel free to remove this figure/sub-experiment, if you have a more interesting figure
% \subsubsection{Level Set Contour Plot and the Normal Space Vector Field of Learned {\ecmnn}}
%
In Fig.~\ref{sfig:levelset_contour_and_jac_vector_field}, we plot the level set contour as well as the normal space eigenvector field of an {\ecmnn} after training on a 3D unit sphere constraint dataset. We see that at both cross-sections $y=0$ (left) and $z=0$ (right), the contours are close to circular, which is expected for a unit sphere constraint manifold.

% 2. compare generated points on the manifold
\subsection{Evaluate Implicit Functions on Generated Configurations}
\label{sec:experiment2}
We use the trained VAE models from Sec. \ref{sec:experiment1} to generate 100,000 new, on-manifold configurations for each constraint. We then evaluate these configurations with the implicit function $h_M$ and report the mean and standard deviation in Table \ref{table:experiment2}.

% 3. compare plans produced by SMP with these methods incorporated 
\subsection{Sequential Motion Planning on Learned Manifolds}
\label{sec:experiment3}
In this experiment, we incorporate a learned manifold into the planning framework developed and introduced in \citet{englert2020sampling}. The Sphere dataset (see Section \ref{sec:datasets}) is used to learn a manifold representation with {\ecmnn}. This learned manifold is combined with two analytical manifolds representing paraboloids. A motion planning problem is defined on these three manifolds where a 3D point has to find a path from a start configuration on one of the paraboloids to a goal configuration on the other. See Figure \ref{sfig:hourglass_learned} for a visualization of the manifolds and a found path with SMP.

\begin{table}
  \caption{Experiment A -- $\constraintfunction$ on Datasets.}
  \label{table:experiment1}
  \centering
  \resizebox{\columnwidth}{!}{%
  \begin{tabular}{ccc|cc|cc|cc|cc|cc}
    \toprule
         Method & \multicolumn{12}{c}{Dataset} \\
     & \multicolumn{2}{c}{Nav} & \multicolumn{2}{c}{Sphere} & \multicolumn{2}{c}{Plane} & \multicolumn{2}{c}{Orient} & \multicolumn{2}{c}{Tilt} & \multicolumn{2}{c}{Handover} \\
     \midrule
              & $\mu$ & $\sigma$ & $\mu$ & $\sigma$ & $\mu$ & $\sigma$ & $\mu$ & $\sigma$ & $\mu$ & $\sigma$ & $\mu$ & $\sigma$ \\
    \midrule
      VAE     & 0.42 & 0.18 & 0.08 & 0.03 & 0.16 & 0.12 & 0.01 & 0.03 & 0.09 & 0.07 & 0.56 & 0.47 \\
    %   \ecmnn     & N/A & N/A & 0.0417 & 0.0583 & 0.45648336 & 1.3869419 & -0.048004348 & 0.9334844 & N/A & N/A & 0.23925124 & 0.7597604 \\
      \ecmnn     & N/A & N/A & 0.04 & 0.06 & 0.46 & 1.39 & -0.05 & 0.93 & N/A & N/A & 0.24 & 0.76 \\
    \bottomrule
  \end{tabular}
  }
\end{table}

\begin{table}
    \caption{Experiment B -- $\constraintfunction$ on Generated Configurations.}
  \label{table:experiment2}
  \centering
  \resizebox{\columnwidth}{!}{%
  \begin{tabular}{ccc|cc|cc|cc|cc|cc}
    \toprule
         Method & \multicolumn{12}{c}{Dataset} \\
     & \multicolumn{2}{c}{Nav} & \multicolumn{2}{c}{Sphere} & \multicolumn{2}{c}{Plane} & \multicolumn{2}{c}{Orient} & \multicolumn{2}{c}{Tilt} & \multicolumn{2}{c}{Handover} \\
     \midrule
              & $\mu$ & $\sigma$ & $\mu$ & $\sigma$ & $\mu$ & $\sigma$ & $\mu$ & $\sigma$ & $\mu$ & $\sigma$ & $\mu$ & $\sigma$ \\
    \midrule
      VAE     & 0.58 & 0.18 & 0.10 & 0.03 & 0.14 & 0.08 & 0.01 & 0.01 & 0.96 & 0.30 & 0.10 & 0.02 \\
    \bottomrule
  \end{tabular}
  }
\end{table}

% \begin{table}
%     \caption{Experiment C -- Motion plans.}   
%   \label{table:experiment3}
%   \centering
%   \resizebox{\columnwidth}{!}{%
%   \begin{tabular}{ccc|cc|cc|cc|cc|cc}
%     \toprule
%          Method & \multicolumn{12}{c}{Dataset} \\
%      & \multicolumn{2}{c}{Nav} & \multicolumn{2}{c}{Sphere} & \multicolumn{2}{c}{3DOF} & \multicolumn{2}{c}{6DOF} & \multicolumn{2}{c}{Tilt} & \multicolumn{2}{c}{Handover} \\
%      \midrule
%               & $\mu$ & $\sigma$ & $\mu$ & $\sigma$ & $\mu$ & $\sigma$ & $\mu$ & $\sigma$ & $\mu$ & $\sigma$ & $\mu$ & $\sigma$ \\
%     \midrule
%       VAE     &  &  &  &  &
%                  &  &  &  &
%                 &  &  & \\
%       \ecmnn     &  &  &  &  &
%                  &  &  &  &
%                 &  &  & \\
%     \bottomrule
%   \end{tabular}
%   }
% \end{table}

% \input{use_in_planning}
\section{Discussion and Future Work}
\label{sec:discussion}
In this paper, we presented ways of learning constraint manifolds for sequential manifold planning. One of them is the novel {\ecmnnlong} ({\ecmnn}). {\ecmnn} is a method for learning representation for implicit functions, with an emphasis on representing equality constraints, while VAEs can also learn inequality constraints. We showed that {\ecmnn} has successfully learned equality constraint manifolds and that these manifolds can be used in a sequential motion planning method.

There are several interesting improvements and future directions to pursue.
First, there are still limitations with the current approach; in particular, our approach does not address the sign/polarity assignments of the implicit function value output, which we plan to address.
Second, we plan to do more comprehensive testing on higher-dimensional manifolds, and incorporate multiple learned constraints into a single motion plan with more subtasks. Further, we also plan to integrate manifolds learned with VAE into motion planning algorithms.

% \section*{Acknowledgments}

%% Use plainnat to work nicely with natbib. 

\bibliographystyle{plainnat}
\bibliography{refs}

\end{document}